\documentclass[final,3p,times,authoryear]{elsarticle}

\usepackage{amssymb}
\usepackage{amsmath}

\usepackage{natbib}
\usepackage{multirow}
\usepackage{array}
\usepackage{longtable}
\usepackage{booktabs}
\usepackage[dvipsnames]{xcolor}
\usepackage{subcaption}
\usepackage{graphicx}

\newcolumntype{L}[1]{>{\raggedright\arraybackslash}p{#1}}
\newcolumntype{C}[1]{>{\centering\arraybackslash}m{#1}}

\newcommand{\edit}[1]{{\color{black}{#1}}}

\DeclareMathOperator{\cair}{CAIR}

\date{}

\journal{Machine Learning with Applications}

\begin{document}

\begin{frontmatter}



\title{Sharing is CAIRing: Characterizing Principles and Assessing Properties of Universal Privacy Evaluation for Synthetic Tabular Data\tnoteref{t1}} 


\author[1,2]{Tobias Hyrup} \ead{hyrup@imada.sdu.dk}
\author[1]{Anton Danholt Lautrup} \ead{lautrup@imada.sdu.dk}
\author[1]{Arthur Zimek} \ead{zimek@imada.sdu.dk}
\author[1]{Peter Schneider-Kamp} \ead{petersk@imada.sdu.dk}

\affiliation[1]{organization={University of Southern Denmark, Department of Mathematics and Computer Science},
            addressline={Campusvej 55}, 
            city={Odense},
            postcode={5230},
            country={Denmark}
            }

\fntext[2]{Corresponding author.}
\tnotetext[t1]{Published in: Hyrup, T., Lautrup, A. D., Zimek, A., \& Schneider-Kamp, P. (2024). Sharing is CAIRing: Characterizing principles and assessing properties of universal privacy evaluation for synthetic tabular data. Machine Learning with Applications, 18, 100608. doi:10.1016/j.mlwa.2024.100608}

\begin{abstract}
Data sharing is a necessity for innovative progress in many domains, especially in healthcare. However, the ability to share data is hindered by regulations protecting the privacy of natural persons. Synthetic tabular data provide a promising solution to address data sharing difficulties but does not inherently guarantee privacy. Still, there is a lack of agreement on appropriate methods for assessing the privacy-preserving capabilities of synthetic data, making it difficult to compare results across studies. To the best of our knowledge, this is the first work to identify properties that constitute good universal privacy evaluation metrics for synthetic tabular data. The goal of universally applicable metrics is to enable comparability across studies and to allow non-technical stakeholders to understand how privacy is protected. 
We identify four principles for the assessment of metrics: Comparability, Applicability, Interpretability, and Representativeness (CAIR). To quantify and rank the degree to which evaluation metrics conform to the CAIR principles, we design a rubric using a scale of 1-4. Each of the four properties is scored on four parameters, yielding 16 total dimensions. We study the applicability and usefulness of the CAIR principles and rubric by assessing a selection of metrics popular in other studies. The results provide granular insights into the strengths and weaknesses of existing metrics that not only rank the metrics but highlight areas of potential improvements. We expect that the CAIR principles will foster agreement among researchers and organizations on which universal privacy evaluation metrics are appropriate for synthetic tabular data.
\end{abstract}



\begin{keyword}
Privacy \sep Privacy evaluation \sep Synthetic data \sep Data sharing \sep Tabular data \sep Machine learning



\end{keyword}

\end{frontmatter}



\section{Introduction} \label{sec:introduction}
The volume of data collected and produced continues to increase as our society moves toward a higher degree of digitalization, and personal and health data are no exception. Regulatory restrictions on data sharing, such as HIPAA~\citep{OfficeforCivilRightsOCR2012GuidanceRule} and GDPR~\citep{EuropeanParliamentandCounciloftheEuropeanUnion2016RegulationRelevance}, have been implemented to protect individuals from privacy violations. The downside of imposing restrictions on data sharing is the substantial limitations on innovation and research based on these data. Consequently, methods that allow data sharing under the regulations are a necessity to catalyze data-sharing capabilities and privacy improvements. Privacy-preserving synthetic data generation (SDG) is a promising attempt to solve this problem by generating simulated records with the same statistical properties as the true data~\citep{Rankin2020ReliabilitySharing, Sun2021AdversarialSurvey}. Synthetic records can be used as a proxy for real data, while allowing analyses to reach highly similar conclusions, thereby maintaining the privacy of individuals. \edit{However, no standardized approach to evaluating the privacy preserving capabilities of synthetic data has been established. Consequently, studies employ a variety of incomparable measures that make cross-study comparisons challenging. This work aims at establishing a foundation on which universally applicable metrics can be developed and improved to consolidate the research community and regulatory units.}

\edit{
\subsection{Privacy Preserving Methods}

Many privacy preserving technologies have been proposed to ensure the protection of individuals. Traditional anonymization techoniques such as k-Anonymity \citep{Samarati1998ProtectingSuppression, Sweeney2002} and $\ell$-diversity \citep{Machanavajjhala2006L-Diversity:K-anonymity} protect privacy by generalizing records to ensure that no unique individuals are present in the data. These methods are simple but have been shown to be vulnerable to adversarial attacks \citep{NarayananNetflix2008, Li_t-closeness_2007}. A different approach is Fully Homomorphic Encryption (FHE) which allows analyses to be conducted on a cipher text rather the original data \citep{Gentry2009AScheme}. However, FHE is computationally expensive, making it impractical for many use cases such as deep learning \citep{joonwoo2022}. A popular method to introduce privacy in algorithms is Differential Privacy (DP), which provides formal privacy guarantees to an algorithm by adding noise to the data or employing noisy operations \citep{Dwork2013ThePrivacy}. Federated learning is a privacy-oriented method to address machine learning problems in which sensitive data are located at multiple clients \citep{ZHANG2021106775}. In a federated learning setting, machine learning models are trained locally and only the model parameters are transferred to a central unit that aggregates models from all clients to a single global model. 

These methods all have their specific use cases and can often be used in combination with each other such as training differentially private models in a federated learning setting. Similarly, synthetic data is useful for specific use-cases and be created using other privacy preserving methods. In cases where the downstream tasks are unknown or where data sharing is needed, synthetic data is a promising solution as an alternative to other privacy preserving methods~\citep{Rankin2020ReliabilitySharing, Sun2021AdversarialSurvey}.
}

\edit{\subsection{Importance of Privacy Evaluation}}
As synthetic data generation is still maturing for tabular data, an abundance of evaluation metrics is applied in various studies. Consequently, comparing the evaluation results of the generated data is a highly challenging task, slowing innovation in generation methods~\citep{Chundawat2022TabSynDex:Data, Ghosheh2022ASources, Hernandez2022SyntheticReview}. Therefore, standardized approaches are needed to evaluate the privacy-preserving capabilities of synthetic tabular data.

In addition, the European Union address potential issues with the application of artificial intelligence (AI) with the Artificial Intelligence Act~\citep{EuropeanComission2021ProposalACTS.} which moves towards a more secure, fair, and private use of AI. Therefore, privacy evaluation is important when dealing with synthetic data. Multiple studies claim that fully synthetic data are inherently private~\citep{Hernandez2022SyntheticReview, Murtaza2023SyntheticDomain} due to the disconnect between true and synthetic data. However, synthetic data have been shown to be successfully attacked and therefore require a thorough privacy evaluation~\citep{ElEmam2020EvaluatingValidation, Hittmeir2020AData}.

\edit{\subsection{Privacy Evaluation}}
Many methods have been proposed to assess the privacy preserving capabilities of synthetic tabular data \citep{breugel2023domias, DAmico2023SyntheticHematology, ElEmam2020EvaluatingValidation, Hu2023CB-GAN:Networks, Rashidian2020SMOOTH-GAN:Generation, Tucker2020GeneratingSoftware, Yale2019PrivacyData, Yoon2020AnonymizationADS-GAN} all of which serve as important steps in ensuring privacy preservation. However, parallel to the development of privacy evaluation metrics is the development of generation methods to establish privacy \citep{Hernandez2022SyntheticReview, Murtaza2023SyntheticDomain}. Nevertheless, when a wide selection of evaluation metrics is available at evaluation time, studies that develop generation methods can select different metrics, causing difficulties in comparing results between works. This effect is exacerbated by evaluation metrics being designed for specific purposes or data types making them inapplicable in some cases.

Although the purpose of privacy evaluation metrics is to accurately gauge the privacy level provided by the generation process, research may be slowed down by having incomparable results across studies. Therefore, there is a need for universal privacy evaluation metrics for synthetic tabular data that can be applied to and compared between multiple studies. Considering the important role regulations have in protecting privacy, we argue that a good universal evaluation metric allows for comparability across works in a way that allows non-technical stakeholders, such as regulators and natural persons, to understand how privacy is protected.

\edit{\subsection{Main Contributions}}
Therefore, the objective of this paper is to define principles for good universal privacy evaluation metrics for synthetically generated tabular data that can aid researchers and regulators in creating and selecting appropriate evaluation methods. We have created a scoring system that uses a rubric to rank existing evaluation metrics on their universality to achieve that goal. A byproduct of establishing universally applicable metrics is that the scoring is not designed to rank metrics by their accuracy, but rather by their versatility across different applications while still maintaining precision in estimating privacy levels.

\edit{The main contributions of our work are the following:}

\edit{
    \begin{itemize}
        \item Characterization of four general principles for universal privacy evaluation methods for tabular synthetic data named CAIR (Comparability, Applicability, Interpretability, and Representativeness). Each principle consists of four dimensions that provide high-fidelity insight into the universality of privacy evaluation metrics.

        \item We provide a scoring rubric for CAIR based on the identified principles that enable users to quantify the degree of conformity to CAIR. The scoring allows granular insights while mitigating bias from qualitatively scoring metrics.
    \end{itemize}
}

In this work, we use the concept of privacy metric and privacy measure synonymously for any function, method, algorithm, or framework that is used to evaluate the privacy level provided by synthetically generating data. To the best of our knowledge, this is the first attempt to define good practices with a scoring system for the evaluation of the privacy of synthetic tabular data.

In the remainder, we first present related work, followed by a presentation of the criteria and rubric developed in this work. Then we detail all dimensions of the rubric and apply them to a selection of existing privacy evaluation metrics. Finally, we discuss the strengths, limitations, and significance of our work followed by a conclusion. As a courtesy for the reader, we supply additional details in \ref{sec:appendix:scores}.

\section{Related Work} \label{sec:related_work}
One of the primary motivations for creating synthetic data is to introduce privacy, but the evaluation of utility and realism has been a primary focus of many researchers~\citep{Baowaly2019RealisticNetworks, breugel2023domias, Chandra2022ImpactsPerformances, Chundawat2022TabSynDex:Data, Dankar2022AGenerators, Farou2020DataGenerator, Ghosheh2022ASources, Hansen2023ReimaginingBenchmark, Murtaza2023SyntheticDomain, Rankin2020ReliabilitySharing}. The same focus on utility is also present when defining universally applicable evaluation metrics, with most research focusing on finding suitable utility or realism metrics~\citep{Chundawat2022TabSynDex:Data, Dankar2022AGenerators, ElEmam2022UtilityStudy, Hernandez2022SyntheticReview, Hernadez2023SyntheticDimensions}.

\edit{\subsection{Privacy Evaluation Approaches}}
Among the many approaches to privacy evaluation are three broad categories \citep{Ganev2023, Murtaza2023SyntheticDomain}: attribute disclosure, membership inference, and similarity-based evaluation. 

Attribute disclosure risk (ADR) is an attack that evaluates an adversary's capability of predicting previously unknown sensitive attributes for individuals, with various levels of adversarial knowledge~\citep{Kaur2021ApplicationData}. 

Membership inference attacks (MIA) aim to infer whether an individual is in the training data. Many variations of MIAs exist ranging from simple attacks based on the proximity of synthetic records to a real record \citep{Yan2020GeneratingConstraints} to more advanced attacks that leverage underlying structures in synthetic and real data such as DOMIAS \citep{breugel2023domias}, LOGAN \citep{Hayes2017LOGANMI}, ReconSyn \citep{Ganev2023}, and privacy gain defined by \citet{stadler2022}. 

Similarity-based metrics use similarity or proximity between synthetic and real records to determine privacy violations. There is a wide range of similarity-based metrics, from the relatively straightforward Distance to Closest Record (DCR) \citep{Guillaudeux2023Patient-centricAnalysis} to more complex metrics such as $\epsilon$-identifiability \citep{Yoon2020AnonymizationADS-GAN} and identity disclosure risk (IDR) by \citet{ElEmam2020EvaluatingValidation}. 

Across all these methods, covering the three categories, there is a shared goal: optimizing the accuracy of privacy evaluation. Although striving for accuracy is crucial, the practical adoption of these metrics faces challenges when they are incompatible with diverse scenarios.

\edit{\subsection{Addressing Comparability Challenges}}
\citet{ElEmam2022UtilityStudy} attempt to quantify how well six different utility metrics can rank the performance of synthetic data generation methods according to a logistic regression prediction. They empirically evaluate 30 different datasets using three SDG methods. The baseline against which they test is the difference between prediction performance for the true and synthetic data using the area under the receiver operating characteristic curve and the area under the precision-recall curve. Using the Page test~\citep{Siegel1988Sciences}, they test whether the utility metrics agree with the baseline. This approach enables empirical quantification of the metrics' performances but is limited to the single scenario of using logistic regression.

\citet{Dankar2022AGenerators} identify four classes of utility metrics for which they select a representative utility metric for testing. Ultimately, they assess whether one of the metrics can be used for overall utility assessment by investigating the correlations between the selected metrics. The authors identify the Hellinger distance as somewhat correlated with two of the three metrics and conclude that it can be used as a standardized utility metric. The study focuses on a limited set of criteria for streamlining evaluation of utility and does not consider elements such as attribute heterogeneity.

\citet{Hernandez2022SyntheticReview, Hernadez2023SyntheticDimensions} identify challenges comparing results across studies. Therefore, they use a proxy scale consisting of ``excellent'', ``good'', and ``poor'' and define the criteria to fall into each level for all included evaluation metrics. The downside of using this scale is the limited granularity for interpretation. Furthermore, not all metrics fit naturally into the three levels. For example, they distribute the result for membership inference attacks (MIAs)~\citep{Sun2021AdversarialSurvey} into ``excellent'' if privacy is preserved and ``good'' otherwise. Thus, even though the method is a step in the right direction for comparability, it provides a limited proxy for utility and privacy evaluation.

The works presented above design methods that emphasize the utility over privacy to compare results across studies. \edit{Furthermore, the approaches have limitations in terms of granularity and applicability, causing the methods to function in a narrow field. Certainly, any attempt at addressing comparability challenges is a step in a constructive direction, but further standardization is required to allow for universally applicable evaluation metrics. Therefore, this work aims to provide a foundation on which universal privacy evaluation metrics can be developed and improved with a common direction. Ideally, such a foundation can consolidate researchers and regulatory units and form agreement for future regulations. Furthermore, addressing these comparability issues can facilitate innovative acceleration by allowing researchers to compare results between works, improving informed decision making when designing new privacy-aware generative approaches.}

\begin{figure}[tb]
  \centering
  \includegraphics[width=.9\textwidth]{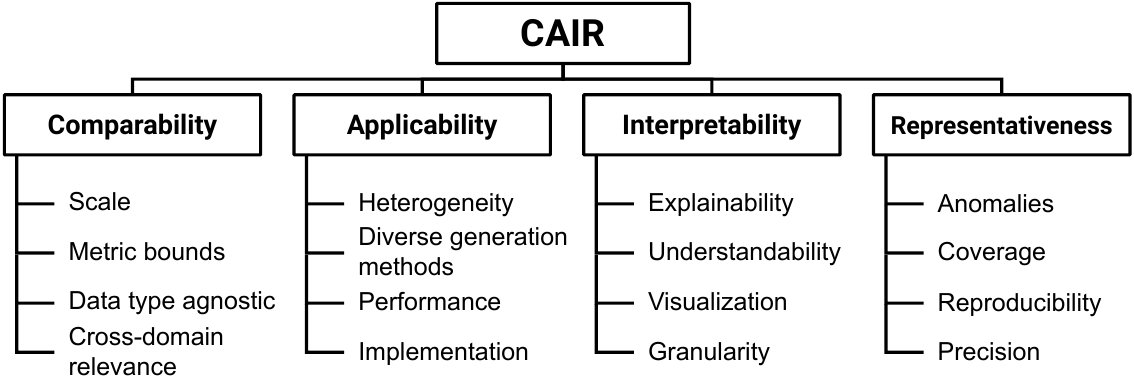}
  \caption{The four CAIR principles with their respective dimensions.}
  \label{fig:cair}
\end{figure}

\section{Method} \label{sec:method}
Assessing the quality of complex work such as privacy evaluation metrics can be a challenging task to perform reproducibly and objectively. Therefore, we break down the assessment into smaller dimensions using a rubric. Rubrics have been successful in educational settings in terms of assessment of lecturer-student, student-lecturer, and peer assessments~\citep{Bowen-Mendoza2022DesignTopics}, but also provide an important assessment tool in other fields of assessment~\citep{Lautrup2023Heart-to-heartAdvice, ODonnell2022QuantitativeAnalysis}. Each dimension should be defined on a scale where the requirements to obtain a specific score are clearly defined to avoid subjectivity and bias.

For this work, four general principles have been identified based on the ideal of a universal privacy evaluation metric and the stakeholders of it:

\noindent\textbf{Comparability:} The ideal privacy metric is comparable across different synthetic datasets.

\noindent\textbf{Applicability:} The ideal privacy metric is applicable to all datasets, data types, and SDG methods.

\noindent\textbf{Interpretability:} The ideal privacy metric can be easily communicated and understood unambiguously by stakeholders.

\noindent\textbf{Representativeness:} The ideal privacy metric accurately represents a realistic privacy level.

The four principles form the acronym CAIR and represent the fact that a universal metric for tabular synthetic data should be usable for all data (Comparability and Applicability) and trustworthy (Interpretability and Representativeness). Each principle is further broken down into four dimensions, which makes the rubric consist of 16 total assessments as illustrated in Figure~\ref{fig:cair}. Maintaining an equal number of unweighted dimensions across all categories ensures uniform weighting among the general categories. The final result is an unweighted mean of all 16 scores. For the sake of accuracy, we allow the inclusion of .5 decimals if a metric is deemed to fall between two scores. The complete rubric is presented in Table~\ref{tab:rubric} using a scale of 1-4 (poor-excellent) with a higher score preferred. Details of each dimension are presented in Section~\ref{sec:cair}.

Metrics can be scored either by a single evaluator or through a multi-evaluator approach. Both settings carry benefits, as the single evaluator setting allows for group discussions and quick estimates of improvement potentials of a metric while the multi-evaluator setting aims at minimizing qualitative biases.

\subsection{Single Evaluator} \label{sec:methods:single-evaluator}
In a single evaluator setting, an individual scores a metric on the 16 CAIR dimensions. \edit{Using this setting is sensitive to potential evaluator bias but serves an important purpose when designing or improving new privacy evaluation metrics. By allowing for single evaluators, researchers can obtain granular insights into strengths and weaknesses of a metric in development without the need to consult independent evaluators.}

The overall CAIR score is given by the mean of all the dimensions. Let $N$ be the number of CAIR dimensions (16 in this work). Let $d_i$ denote the $i^{th}$ dimensions such that $d_i \in [1, 1.5, \dots, 4]$. Then, the CAIR score for a single evaluator is given by:
\begin{equation} \label{eq:single_cair}
    \cair_{single} = \frac{1}{N} \sum_{i=1}^{N} d_i.
\end{equation}

To encourage discussions during metric development, a group of people can collectively assume the role of a single evaluator and report a single score for each dimension.

\subsection{Multiple Evaluators} \label{sec:methods:multi-evaluator}
\begin{figure}[tb]
    \centering
    \includegraphics[width=.9\textwidth]{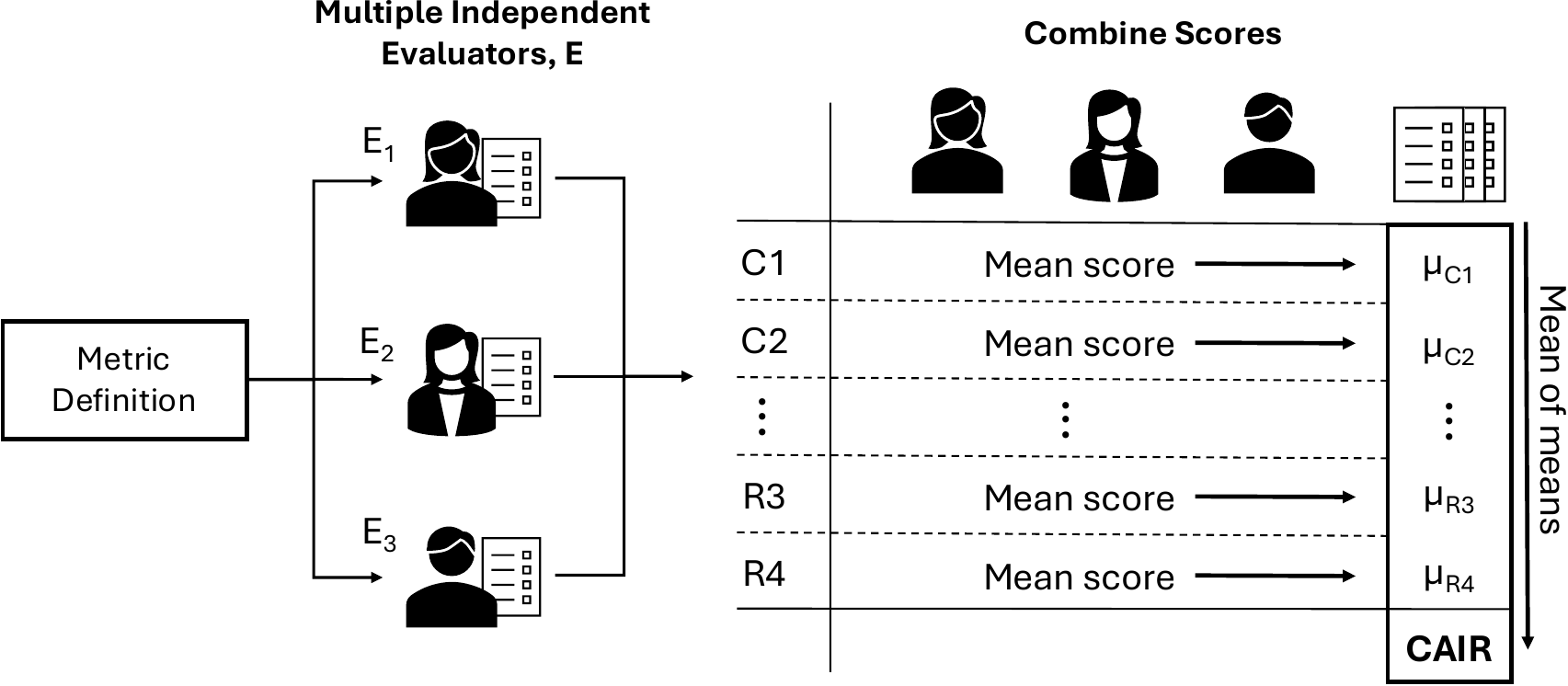}
    \caption{Visual representation of the multi-evaluator CAIR evaluation workflow defined in Equation~\ref{eq:multi_cair}. Given the definition of a privacy metric, a series of independent evaluators score the metric on the CAIR dimensions. For each dimension, the mean across the evaluators' scores are computed with the final CAIR score being the average of dimension means.}
    \label{fig:cair_evaluation_flow}
\end{figure}

To mitigate qualitative bias by individual evaluators, multiple independent evaluators can score each metric on the 16 dimensions. Contrary to the single-evaluator setting of a group, this setting requires all evaluators to individually score the given metric. The standard error is used as the uncertainty estimate based on the assumption that by increasing the number of evaluators, the CAIR score approaches a consensus. Consequently, the error must be propagated to the final CAIR score. The evaluation workflow is depicted in Figure~\ref{fig:cair_evaluation_flow} and is defined below.

To account for multiple evaluators, we extend Equation~\ref{eq:single_cair} accordingly. Let $N$ be the number of CAIR dimensions. Let $\mathcal{E}$ be a domain of individual independent evaluators $E \in \mathcal{E}$, i.e., functions such that \mbox{$E : d_i \rightarrow \{1, 1.5, \dots, 4\}$}, $i=1 \dots N$ where $d_i$ is the corresponding CAIR dimension. Then, the CAIR score is \edit{defined as:}
\begin{equation} \label{eq:multi_cair}
    \cair = \frac{1}{N} \sum_{i=1}^{N} \frac{1}{|\mathcal{E}|} \sum_{E \in \mathcal{E}} E(d_i).
\end{equation}

As the number of evaluators grows, the CAIR score approaches a consensus as defined by the increasing precision due to the shrinking standard error, $s_{\overline{\mathit{CAIR}}}$ (propagated) \edit{defined in Equation~\ref{eq:prop_error}. A CAIR score computed using the multi-evaluator setting should be reported with the propagated error to quantify dimension-wise discrepancies across evaluators. A higher error indicates that evaluators exhibit a higher degree of disagreement, indicating that more evaluators are needed and that the score should be interpreted with care. Considering that the multi-evaluator setting mitigates bias by using independent evaluators and reports on their disagreements, the setting is suitable for comparing multiple metrics to make informed decisions on model performance evaluation.}

\begin{equation} \label{eq:prop_error}
    s_{\overline{\mathit{CAIR}}} = \frac{1}{N}\sqrt{\sum_{i=1}^{N} s_{\overline{d_i}}^2}.
\end{equation}

\begin{figure}
    \centering
    \begin{subfigure}[b]{0.22\textwidth}
        \includegraphics[width=\textwidth]{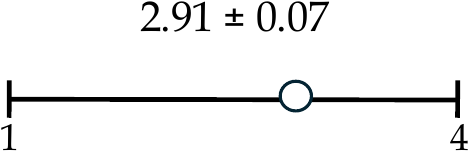}
        \vspace{20pt}
        \caption{\edit{CAIR Score}}
        \label{fig:granularity:cair_score}
    \end{subfigure}
    \hfill
    \begin{subfigure}[b]{0.4\textwidth}
        \includegraphics[width=\textwidth]{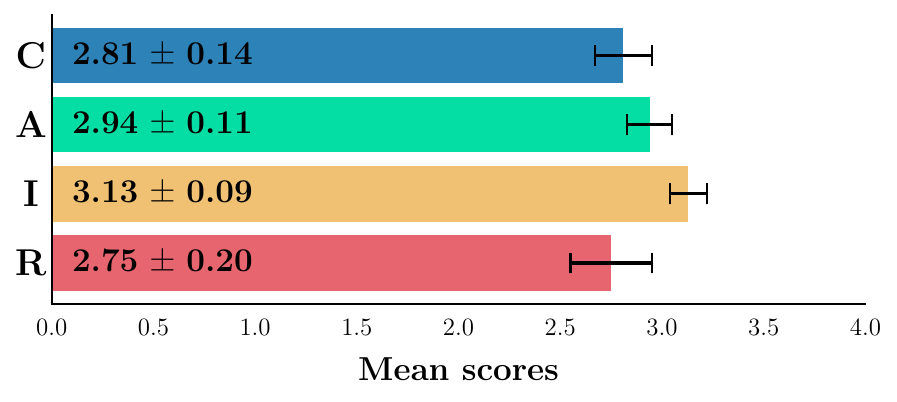}
        \caption{\edit{Grouped by Principles}}
        \label{fig:granularity:principles}
    \end{subfigure}
    \hfill
    \begin{subfigure}[b]{0.22\textwidth}
        \includegraphics[width=\textwidth]{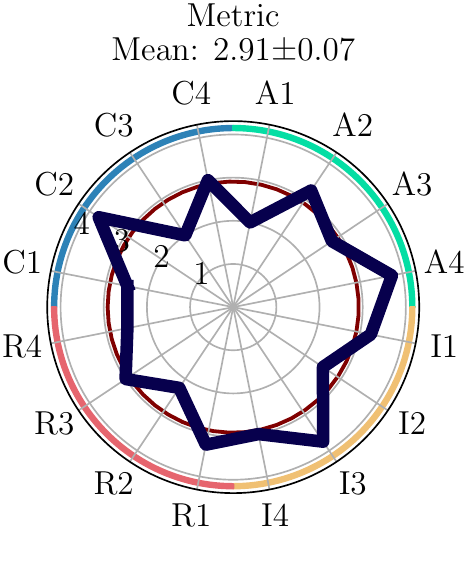}
        \caption{\edit{Individual Dimensions}}
        \label{fig:granularity:dimension-wise}
    \end{subfigure}
    \caption{\edit{An illustrative example of CAIR assessments displayed with varying levels of granularity. (a) displays the general CAIR score as an aggregate value. (b) presents the scores grouped by the corresponding principle. (c) illustrates scores on individual dimensions.}}
    \label{fig:granularity}
\end{figure}

\edit{
\subsection{Levels of Granularity}
Segmenting the principles into multiple dimensions allows for varying levels of granularity when examining the assessments. Figure~\ref{fig:granularity} presents three levels of granularity, each with different benefits. The overall CAIR score in Figure~\ref{fig:granularity:cair_score} is the most general representation, being an aggregate of all dimensions across all evaluators. It provides a single value, that is easy to compare between metrics. Figure~\ref{fig:granularity:principles} groups the scores according to the respective principles and provides internal propagated errors. This representation allows differentiation in performance across principles rather than over all dimensions. The most detailed representation is presented in Figure~\ref{fig:granularity:dimension-wise} where the mean score for each dimension across evaluators. This representation enables a detailed investigation into strengths and weaknesses. For the most comprehensive investigation, the dimension-wise representation in Figure~\ref{fig:granularity:dimension-wise} is recommended.
}

\section{The CAIR Principles}\label{sec:cair}
The CAIR principles for good privacy evaluation for synthetic tabular data consist of a range of criteria that aim to describe the ideal privacy measure. The rubric in Table~\ref{tab:rubric} is limited in description depth, but should be sufficient to score any privacy metric without additional information. However, more detailed information on the dimensions is necessary to adequately argue for their inclusion.

\subsection{Comparability} \label{sec:comparability}
One of the primary motivations for advocating standardized privacy evaluation metrics lies in the concept of comparability. By establishing a consistent set of metrics, researchers and practitioners can systematically compare privacy outcomes across various models, datasets, and projects. This comparability not only facilitates informed decision-making but also fosters healthy competition within the research community of synthetic data generation. With the ability to assess the efficacy of different generation methods, stakeholders can identify the most effective strategies for preserving privacy. Comparisons accelerate innovation by allowing results and generation methods to be compared and shared between domains and context-specific applications.

\subsubsection*{C1 Scale} \label{sec:comparability:scale}
Comparisons of privacy levels between generation processes rely on the privacy measure being on the same scale. If the scale is not identical, the resulting evaluation values are incompatible, making any attempt to compare the results futile. An argument can be made that users should normalize the data before computing the evaluation results. However, considering that synthetically generated data are primarily useful on the same scale as the true data, it is unlikely that all users will report the results on normalized data. In addition, the scale of the data can skew the results of a measure even if the metric has well-defined bounds. Consider the function \mbox{$f(\mathcal{T}, \mathcal{S}) = \sigma(\lVert \mathcal{T}_i - \mathcal{S}_j \rVert_2)$}, where \(\mathcal{T}\) and \(\mathcal{S}\) are true and synthetic data respectively and \(\sigma\) is the sigmoid function; then if two attributes defined as $a_0 \in (0, 10^6)$ and $a_1 \in (0,1)$ are in the dataset, the influence of $a_0$ on the function will be substantially greater than that of $a_1$ even though the final result is in $(0, 1)$.

\subsubsection*{C2 Metric Bounds} \label{sec:comparability:metric_bounds}
Well-defined upper and lower bounds of a metric are important for accurately comparing evaluation results. Without fixed bounds, the results from one dataset or generation method to another do not correspond, even if the value is the same. Whether the bounds are (0, 1), (-1, 1), or something else is not important, as long as the bounds are consistent.

\subsubsection*{C3 Data Type Agnostic} \label{sec:comparability:data_type_agnostic}
Heterogeneous data types are prevalent in tabular data~\citep{Murtaza2023SyntheticDomain}. As such, it is crucial that the heterogeneity of the data has a limited effect on the output of the metric. Ideally, the metric should be invariant to any tabular data type, but a remediation can be to perform appropriate pre-processing. If pre-processing cannot mitigate adverse effects of the mixed data types, further analysis should be conducted to identify whether a systematic bias exists towards any specific subsets of data types. Consider the use of Hamming distance to determine the similarity between synthetic and true data. Hamming distance computes the number of attributes where the records differ and as such performs best for categorical data. Some pre-processing such as data binning can mitigate parts of the inaccuracies on continuous data, but it can potentially have significant adverse effects on the results.

\subsubsection*{C4 Cross-Domain Relevance} \label{sec:comparability:cross-domain_relevance}
Privacy-aware synthetic data generation is relevant in a wide variety of domains. Therefore, it is no surprise that many generation methods are designed specifically for their work domains to increase utility and privacy. However, methods explicitly applied to a niche domain might be useful for research in unrelated domains. Cross-domain comparisons and interpretations can be complex and cumbersome if the privacy evaluation method applied in a domain-specific work is equally niche. Therefore, a comparable evaluation method must work in various domains. Nevertheless, specialized metrics are by no means obsolete, as they can provide a unique perspective on particular problems. However, they should not be applied in a general scenario unless they are accompanied by a universal metric that allows for comparability.

\begin{scriptsize}
\begin{longtable}{C{0.05\textwidth}L{0.21\textwidth}L{0.21\textwidth}L{0.21\textwidth}L{0.21\textwidth}}
    \caption{Rubric consisting of four categories (Comparability, Applicability, Interpretability, Representativeness), each with four dimensions. Scoring is performed on a scale of 1-4 with 0.5 decimals allowed when a metric falls between two scores.} \label{tab:rubric} \\
    \toprule
    & \textbf{Excellent 4} & \textbf{Good 3} & \textbf{Fair 2} & \textbf{Poor 1}  \\ \midrule
    \endfirsthead
        \toprule
         & \textbf{Excellent 1} & \textbf{Good 3} & \textbf{Fair 2} & \textbf{Poor 1}  \\ \midrule
         \endhead
         \\
         \multicolumn{5}{l}{\normalsize \textbf{Comparability}} \\[1em]
         \multirow{1}{*}[-2em]{\hfill \rotatebox[origin=c]{90}{\textbf{Scale}}} & 
            - Completely invariant to the unit scale of attributes. \newline - The range of the metric is not affected by the scale of the data. &
            - Slightly sensitive to attributes having different scales. \newline - The range of the metric is not affected by the scale of the data. &
            - Somewhat sensitive to attributes having different unit scales. \newline - The scale of the metric is relative to the scale of the data. & 
            - The metric is highly sensitive to attributes having differences in unit scales. \newline - The scale of the metric is relative to the scale of the data. \\ \\ \cmidrule{2-5}
        \\
        \multirow{1}{*}{\rotatebox[origin=c]{90}{\parbox{2cm}{\centering \textbf{Metric \\ bounds}}}} & 
            - The metric has a clearly defined upper and lower bound. &
            - The metric has an upper and lower bound. \newline - The bounds are defined, but they change relative to the data, e.g., can be (0, 1) for one dataset and (1, 50) for another. &
            - The metric has either an upper or lower bound but not both. & 
            - The metric has no upper or lower bounds. \\ \\ \cmidrule{2-5}

        \\
        \multirow{1}{*}{\rotatebox[origin=c]{90}{\parbox{1.8cm}{\centering \textbf{Data type \\ agnostic}}}} & 
            - Mixed data types have little to no influence on the analysis and do not significantly affect results or conclusions. & 
            - Mixed data types introduce mild inaccuracies in the output. \newline - Issues can be managed through appropriate data pre-processing or analysis techniques. & 
            - Some combinations of mixed data types have significantly adverse effects on the analysis. \newline - Remediations require specialized techniques or complex efforts to handle properly. &
            - Mixed data types severely undermine the validity and reliability of the results, making the analysis highly complicated or infeasible. \\ \\ \cmidrule{2-5}
            
        \\
        \multirow{1}{*}{\rotatebox[origin=c]{90}{\parbox{1.5cm}{\centering \textbf{Cross-domain relevance}}}} & 
            - The metric shows comprehensive relevance across diverse domains. \newline - Accommodates privacy and domain challenges effectively. & 
            - The metric shows moderate cross-domain relevance, but may not cover all domains equally or comprehensively. &
            - The metric has limited relevance across different domains. \newline - Useful for more than just a narrow domain. &
            - The metric is highly domain-specific. \newline - Lacks relevance outside of a narrow domain. \\\\\\\\
        
        \midrule
        \\
        \multicolumn{5}{l}{\normalsize \textbf{Applicability}} \\[1em]
         \multirow{1}{*}[-0.5em]{\hfill \rotatebox[origin=c]{90}{\textbf{Heterogeneity}}} & 
            - Applicable to most tabular data types. \newline - Can handle mixed tabular data types in the same dataset. \newline - Has no bias towards certain data types. &
            - Applicable to most tabular data types. \newline - Can be modified to accommodate mixed data types, but no definition is provided. \newline - Has a slight bias towards certain data types. &
            - Applicable to at least two different data types. \newline - Struggles with handling mixed data types with limited adaptability. \newline - Has some bias towards certain data type(s). & 
            - Applicable to only one data type. \\ \cmidrule{2-5}

        \\
        \multirow{1}{*}{\rotatebox[origin=c]{90}{\parbox{1.2cm}{\centering \textbf{Diverse generation methods}}}} & 
            - Applicable to all synthetic data sets regardless of generation method. \newline - No bias towards any methods. &
            - Applicable to more than one family of generative methods. \newline - Has some bias towards some methods. &
            - Is only defined for synthetic data generated using a single family of generative methods, e.g., GANs or CARTs. & 
            - Can only be applied to synthetic data generated with one specific generative method. \\ \\ \cmidrule{2-5}
            
        \\
        \multirow{1}{*}{\rotatebox[origin=c]{90}{\parbox{1.25cm}{\centering \edit{\textbf{Performance}}}}} & 
            - The required runtime is reasonable. $\mathcal{O}(n^2)$. & 
            - The metric balances accuracy and runtime, but further improvements to efficiency can be accomplished. $\mathcal{O}(m n^2)$ where $1 < m \ll n$. & 
            - The metric is computationally expensive. $\mathcal{O}(n^3)$.&
            - Requires substantial computational resources. $\Omega(n^3).$\\\\ \cmidrule{2-5}

        \\
        \multirow{1}{*}[0em]{\hfill \rotatebox[origin=c]{90}{\textbf{Implementation}}} & 
            - Easy to implement and/or an implementation is supplied. \newline - The metric is computed using only readily accessible information. & 
            - Definition or documentation of the metric has a few ambiguous elements that make consistent implementation somewhat challenging. \newline - The data required to compute the metric exist in the dataset and do not require information from external sources such as statistics or datasets for comparisons. &
            - The metric poses ambiguity and some practical challenges making it difficult to apply in real-world scenarios. \newline - The required information might be challenging to obtain from external sources and/or to compute. &
            - The metric is poorly documented and contains multiple elements open for interpretation. \newline - The metric is very impractical to implement, and no implementation is provided. \newline - The required information is very difficult or impossible to obtain from external sources and/or compute. \\ \\
        
        \midrule
        \multicolumn{5}{l}{\normalsize \textbf{Interpretability}} \\[1em]
         \multirow{1}{*}[-1em]{\hfill \rotatebox[origin=c]{90}{\textbf{Explainability}}} & 
            - It is easy to communicate what the metric measures. \newline - The metric is straightforward and minimizes unnecessary complexity. &
            - The metric has some complex elements that may be difficult to communicate to a layperson in terms of what is measured. \newline - Computation includes unintuitive or complex elements that can be challenging to communicate to a layperson. &
            - The metric is complex to the degree that it is unlikely that the metric can be explained to a layperson. \newline - The metric requires some domain knowledge to comprehend. &
            - The metric is highly complex and is challenging to communicate to people with a technical background. \\ \cmidrule{2-5}

        \multirow{1}{*}[-0.5em]{\hfill \rotatebox[origin=c]{90}{\textbf{Understandability}}} & 
            - The output of the metric is easy to understand for a layperson without the need for explanations. &
            - The output of the metric can be understood by a layperson, but might require some explanation. &
            - Understanding the output of the metric requires some background knowledge and is somewhat difficult to communicate to a layperson. \newline - The metric contains some unnecessary complexity that could be simplified. & 
            - The output of the metric requires substantial technical background knowledge to understand and is highly challenging to communicate to a layperson. \newline - The metric is overly complex and difficult to grasp. \\ \cmidrule{2-5}
            
        \multirow{1}{*}[-1.2em]{\hfill \rotatebox[origin=c]{90}{\textbf{Visualization}}} & 
            - The metric can easily be visualized both in isolation and compared to the results of other datasets. \newline - Visualizations enhance the ability to accurately compare multiple results. & 
            - The metric can easily be visualized in isolation. \newline - Making comparative visualizations of multiple results in a single graph can be difficult, but can enhance the ability to differentiate between results. & 
            - The metric can be visualized, but only in isolation. \newline - Comparisons with the results of other datasets are highly difficult or impossible to produce in a single or multiple graphs. &
            - The metric is not suited for visualization. \\ \cmidrule{2-5}
        
        \multirow{1}{*}[-2em]{\hfill \rotatebox[origin=c]{90}{\textbf{Granularity}}} & 
            - The metric offers fine-grained granularity that allows for unambiguous differentiation between privacy levels. \newline - Consists of a single value that makes sense in isolation. & 
            - The metric provides a moderate level of granularity, but some important distinctions might be overlooked. \newline - Consists of a single value that makes sense in isolation. &
            - The metric has limited granularity, making it difficult to distinguish between privacy levels. \newline - Consists of a single value, but only makes sense when compared to other datasets or generation methods. &
           - The metric lacks the ability to clearly differentiate between privacy levels. \newline - Can consist of multiple values that describe different privacy aspects. \newline - Might only make sense in comparison to the privacy level of other datasets, e.g., a holdout set. \\

            \midrule \\

        \multicolumn{5}{l}{\normalsize \textbf{Representativeness}} \\[1em]
         \multirow{1}{*}[0em]{\hfill \rotatebox[origin=c]{90}{\textbf{Anomalies}}} & 
         - The sensitivity of all rare observations are taken into account. \newline - Computation of the metric includes all observations with appropriate weights. &
        - Rare observations are treated with greater weight than common observations. \newline - Computation of the metric prioritizes rare observations. &
        - Rare observations are included in computing the metric. \newline - Rare and common observations are treated equally. &
        - Rare observations are not considered when computing the metric. \\ \\ \cmidrule{2-5}

        \\
        \multirow{1}{*}[0em]{\hfill \rotatebox[origin=c]{90}{\textbf{Coverage}}} &
        - The privacy result is equally accurate for all records. &
        - The privacy result is approximately equally accurate for all records without any systematic bias. &
        - The privacy result is representative for some proper subsets of the data with a systematic bias. &
        - The privacy result only represents a small proper subset of the data. \\ \\ \cmidrule{2-5}

        \\
        \multirow{1}{*}[-0.8em]{\hfill \rotatebox[origin=c]{90}{\textbf{Reproducibility}}} & 
        - The metric always returns the same value for the same input data. &
        - The metric returns the same value for the same input data with a probability close to 1 when applied by different people.  &
        - The metric returns almost the same value for the same input data with some probability, but the variance is somewhat consistent when applied by different people. \newline - An error is provided for the output. &
        - The metric returns significantly different results for the same input. \\ \\ \cmidrule{2-5}

        \\
        \multirow{1}{*}[0em]{\hfill \rotatebox[origin=c]{90}{\textbf{Precision}}} & 
        - The metric does not have systematic over- or underestimation of the privacy level. &
        - The metric slightly underestimates the privacy level but never overestimates it. &
        - The metric either over- or underestimates the privacy level, but not systematically. &
        - The metric significantly over- or underestimates the privacy level, systematically. \\ \\
        \bottomrule
        
\end{longtable}
\end{scriptsize}

\subsection{Applicability} \label{sec:applicability}
The efficacy of a universal privacy evaluation metric is highly dependent on its applicability across diverse datasets and data generation methods. A universally applicable metric must have the ability to assess the privacy-preserving capabilities of synthetically generated tabular data comprehensively, irrespective of the generation techniques employed or the data characteristics. Furthermore, it should be feasible to compute the metric using readily available information and computational resources. This aspect of applicability ensures that the metric remains practical and accessible, allowing stakeholders to evaluate privacy effectively across a broad range of synthetic datasets without significant computational challenges or data-specific constraints.

\subsubsection*{A1 Heterogeneity} \label{sec:applicability:heterogeneity}
The heterogeneous nature of tabular data can pose a challenge when evaluating privacy. Many evaluation methods rely on a measure of similarity between synthetic and true records. However, many popular similarity functions, such as Euclidean and Hamming distance, perform poorly with more than one specific data type. Commonly, authors declare that any similarity function can be applied and rely on the reader to implement the metric with necessary adjustments~\citep{ElEmam2020EvaluatingValidation, Yoon2020AnonymizationADS-GAN}. Proposing metrics with vague definitions restricts the ability to compare results across various implementations accurately. Accordingly, a good privacy evaluation metric considers how data with heterogeneous data types should be handled.

\subsubsection*{A2 Diverse Generation Methods} \label{sec:applicability:diverse_generation_method}
The development and identification of generation methods and privacy implementations that balance utility and privacy is an important aspect of privacy-aware synthetic data research. Therefore, it is crucial that a universal privacy metric is applicable to diverse generation methods. Parametric generation methods imprint a certain characteristic on the synthetic data, which some evaluation methods can potentially favor. Furthermore, evaluation methods that are run during model training or fitting and are dependent on the iterative nature of such processes are likely to favor the methods for which they are designed. These evaluation methods negatively affect the applicability and, therefore, the comparability of privacy metrics.

\subsubsection*{\edit{A3 Performance}} \label{sec:applicability:performance}
\edit{Computational efficiency is crucial when evaluating the practicality of a method, especially when considering the scalability to larger datasets. A method may provide accurate or granular results, but if it is computationally expensive, its applicability diminishes. Many evaluation methods for synthetic data are based on pairwise distances, which involves comparing every point in the true dataset \(\mathcal{T}\) with every point in the synthetic dataset~\(\mathcal{S}\) \citep{Hernandez2022SyntheticReview, Murtaza2023SyntheticDomain}. In this context, the performance of a metric should ideally not exceed \(\mathcal{O}(n^2)\) where $n = \max (|\mathcal{T}|, |\mathcal{S}|)$. While \(\mathcal{O}(n^2)\) is not the most efficient time complexity, it is the worst case for pairwise comparisons, thus defining a natural threshold for performance.}

\subsubsection*{A4 Implementation} \label{sec:applicability:implementation}
A privacy evaluation method is only useful if it is realistic to implement and compute. Ideally, a well-documented implementation of the evaluation metric should be provided whenever a new method is proposed. At the very least, the method should be well-documented to allow for easy, consistent implementation by the reader and to avoid any ambiguity. Consequently, a measure will gain a low implementation score if it contains many complex elements that require extensive domain knowledge to implement appropriately. Furthermore, the data required to compute the metric should be readily accessible, without external data being required. We distinguish between internal and external information: Internal information is contained in the datasets for which the metric is computed, and external information is gathered elsewhere. An example of this property is illustrated in the application of CAIR later in the paper.

External information includes, but is not limited to, dynamic statistics, additional datasets for comparisons or similar, and evaluation results beyond the dataset that is currently being evaluated.

\subsection{Interpretability} \label{sec:interpretability}
A fundamental objective of preserving privacy in synthetic data is to enable data sharing without compromising individual privacy rights. Ultimately, regulatory bodies define what constitutes adequate privacy levels, making them pivotal stakeholders to privacy preservation. The \textit{Interpretability} principle of CAIR governs stakeholders' ability to comprehend how privacy levels are measured and, ultimately, how data privacy is established through data synthesis. Therefore, it is crucial that the privacy levels determined by evaluation metrics can be effectively communicated and understood by the relevant stakeholders, including data regulators and organizational decision-makers.

\subsubsection*{I1 Explainability} \label{sec:interpretability:explainable}
What is measured and how it evaluates privacy should be easy to explain to any stakeholder. Using an explainable metric increases the transparency and, thereby, improves the confidence in the metric. \edit{Explainability can be understood in multiple ways including using easy-to-understand concepts. Relying on complex formulas and black-box models can decrease the explainability to non-technical stakeholders, thus creating barriers for adopting the metric in regulatory decisions. To mitigate the perceived complexity of an evaluation metric, clear formulations and intuitive descriptions can be provided with non-technical stakeholders in mind. A metric that includes elements that can only be understood with technical knowledge inhibits comprehension. Therefore,} explainability increases the probability that the privacy provided by synthetic data is recognized and accepted by any relevant stakeholder.

\subsubsection*{I2 Understandability} \label{sec:interpretability:understandable}
An understandable metric is easy for laypeople to interpret without any additional explanation. That is, the output of the metric is understandable for laypeople. Naturally, a number without context is meaningless. Therefore, the criterion for an understandable metric is that it can be phrased in such a way that no ambiguity is present and no elaboration is necessary. An example could be ``the risk of re-identification is 9\%'', which would be more straightforward to interpret than something such as the evaluation of an adversarial prediction model ``recall of 0.2 and precision of 0.3''.

\subsubsection*{I3 Visualization}
Visualization is an important tool for improved understanding and reduced cognitive load. It allows people to see the data not only in isolation, but in comparison to one another. Therefore, a natural addition to \textit{Interpretability} is visualization. Naturally, metrics that output single values can be visualized both in isolation and in comparisons, but so can other types of output whether it is multiple values, a distribution, or a range. Having visualization as a dimension allows metrics that produce multi-valued outputs to also receive a decent \textit{Interpretability} score as long as they can be interpreted with the right means.

\subsubsection*{I4 Granularity} \label{sec:interpretability:granularity}
Granularity refers to a user's ability to unambiguously differentiate between privacy levels. The most basic form of fine-grained differentiation is to ensure that the privacy measure follows a monotonic function. Any non-monotonic function can confuse a user, especially laypeople. Furthermore, a linear function is preferred as the privacy levels change at a constant rate as a function of the underlying computation. Consequently, any privacy metric that forces well-defined bounds by transformation, for example, by using a sigmoid or hyperbolic tangent function, will be slightly penalized. In addition, the level of granularity decreases if the output of the metric consists of more than a single value such as precision and recall for an adversarial membership inference attack~\citep{Sun2023GeneratingPrivacy}. Keeping track of more than one value means that the user must understand and interpret the individual outputs and how they change in connection. Lastly, the privacy level should be understandable in isolation. The results of some measures may require comparisons with the results of other datasets or generation methods to accurately differentiate between privacy levels. Such metrics should be penalized for this need.

\subsection{Representativeness} \label{sec:representative}
The concept of \textit{Representativeness} captures the ability of privacy metrics to accurately gauge privacy levels of synthetic datasets. An effective metric must possess the capacity to comprehensively capture privacy violations and quantify the severity of these violations. Essentially, it should reflect the complexity of real-world privacy concerns while equally representing the privacy levels of all individuals. Failure to sufficiently capture privacy violations undermines the validity of the metric. Thus, the \textit{Representativeness} principle is an essential part of assessing the efficacy of privacy evaluation metrics for synthetic data.

\subsubsection*{R1 Anomalies} \label{sec:representative:anomalies}
Anomalous records typically have a higher risk of being subject to privacy violations~\citep{Yoon2020AnonymizationADS-GAN}. Inliers typically have the benefit of ``blending in'' with other common observations, ensuring that even highly similar synthetic records provide sufficient privacy levels. Outliers do not enjoy the same security. Synthetic records in the immediate proximity of a true record are likely to be uniquely linked to each other, constituting a privacy violation. Accordingly, anomalous observations should be given appropriate considerations when evaluating the overall level of privacy. Considering that many methods use some form of similarity measure, methods can be constructed to include weighted versions of the similarity function with the weights representing the degree to which individual records are outliers. Weighting of anomalous records results in records close to an outlier seeming more similar to the outlier than is the case when using an unweighted distance function.

\subsubsection*{R2 Coverage} \label{sec:representative:coverage}
The privacy level reported by any metric should be accurate for all the records that the metric claims to include. Accordingly, if the privacy result is only accurate for some subsets of the data, the privacy measure does not represent the overall privacy level. One of the most obvious traps is to give too much weight to anomalous observations, thereby making the privacy level accurate only for outliers or even a subset of the outliers. This restriction is intended to create a balance between R1 and R2 such that it is not possible to obtain an inflated score by assigning excessive weights to outliers.

\subsubsection*{R3 Reproducibility} \label{sec:representative:reproducability}
Metrics that provide different results when applied by different people or at different times carry some form of uncertainty about the metric's accuracy. Therefore, an ideal privacy metric produces the same result when the input is the same. Most metrics that rely on some form of randomization will inevitably produce different results. Such metrics include, but are not limited to, prediction methods that require a random initialization state and methods that evaluate randomized batches of the data. Nevertheless, results can be somewhat consistent even with randomized components and, in such cases, the error should be clearly stated. Including an error does not constitute returning an extra value as part of the \textit{I4 Granularity} dimension.

\subsubsection*{R4 Precision} \label{sec:representative:precision}
A good evaluation metric neither over- nor underestimates the level of privacy provided by the synthetic data. This property is referred to as the precision of the metric. Overestimating privacy levels can be detrimental to the individuals contained in the data by providing a false sense of security. On the other hand, underestimation provides a conservative estimate of the privacy level, which does not harm individuals, but it does constitute an obstacle in optimizing the utility-privacy trade-off.

\section{Application of CAIR} \label{sec:application}
\begin{figure}[t]
    \centering
    \includegraphics[width=0.9\textwidth]{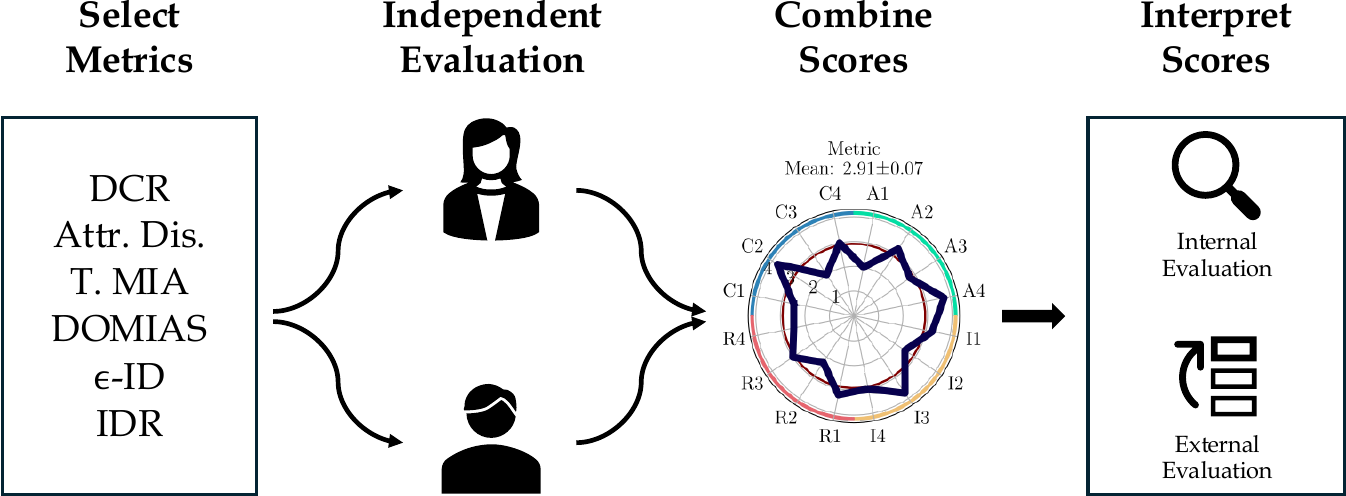}
    \caption{\edit{Workflow for the application setup, demonstrating how CAIR is used and interpreted. A selection of metrics is assessed by two independent evaluators. Their scores are combined to dimension-wise means. Interpreting the results are conducted in two parts: 1) internally where the dimension-wise scores are assessed and 2) externally, where the metric assessments are compared to each other. Metric acronyms: DCR (Distance to Closest Record); Attr. Dis. (Attribute Disclosure); T. MIA (Threshold-based Membership Inference Attack); $\epsilon$-ID ($\epsilon$-Identifiability); IDR (Identity Disclosure Risk).}}
    \label{fig:application_flow}
\end{figure}

\edit{
Figure~\ref{fig:application_flow} illustrates the workflow of the demonstration in this section. First, a diverse selection of privacy evaluation metrics is identified. Two independent evaluators score the metrics on the 16 dimensions of CAIR. Aggregating their scores according to Figure~\ref{fig:cair_evaluation_flow} allows for a dimension-wise representation for a comprehensive analysis. Lastly, the assessments are interpreted in two ways: internally and externally. Internally, the strengths and weaknesses of each metric are identified, allowing high-fidelity insights that can lead to future improvements development strategies. External evaluation of the metrics involves comparing the results between all the metrics, both in terms of the individual dimensions and the overall CAIR score that allows for a ranking of the metrics based on their universality.
}

The results are depicted in Figure~\ref{fig:cair_scores_radar} with the overall score and error being computed using Equations \ref{eq:multi_cair} and \ref{eq:prop_error}, respectively. The scores provided in this section serve as an illustrative example rather than a definitive assessment. This section covers how CAIR can be used to identify areas of improvement for metrics and to gauge what metrics best conform to the concept of universality. Furthermore, the resulting ranking reflects the versatility of the privacy evaluation metrics, rather than solely their ability to precisely gauge the level of privacy. \edit{For a comprehensive analysis of each metric's performance, the dimension-wise representation in Figure~\ref{fig:granularity:dimension-wise} is selected as the appropriate level of granularity.} Detailed information on the scorings is provided in \ref{sec:appendix:scores}.

\subsection{Metrics for Evaluation}
\edit{
To ensure a comprehensive demonstration of the capabilities of CAIR, a diverse selection of metrics is required. Privacy metrics for synthetic tabular data can be divided into three groups: Membership inference attacks (MIA), Attribute disclosure Risk, and similarity-based measures \citep{Murtaza2023SyntheticDomain, Ganev2023}. For each of these categories, a wide variety of metrics have been proposed. For this demonstration, we select metrics from all three categories, ranging from simple naïve to more complex measures.
}

\textit{Distance to the closest record (DCR)} is a simple metric where the output is typically the mean or median minimum distance from a synthetic record to a true one~\citep{Guillaudeux2023Patient-centricAnalysis}. In the most basic scenario, the DCR is computed using Euclidean distance. DCR can be used as a metric on its own, or it can be extended to membership inference.

\textit{Attribute disclosure attacks (Attr.~disclosure)} are similar to MIAs, but instead of inferring membership, the adversary attempts to predict unknown attribute values~\citep{Kaur2021ApplicationData}. An adversary is assumed to have access to a dataset similar to the synthetic data but with at least one attribute missing. The adversary can use a $k$-nearest neighbor algorithm, and the majority vote predicts the univariate values for $k>1$. The attack is typically performed on various levels of prior adversarial knowledge and the output is all of the results or aggregate statistics. Similar to threshold-based membership inference, the reported adversarial performance is the mean precision and recall over all levels of prior knowledge.

\textit{Threshold-based membership inference attacks (T.~MIA)} are attacks in which an adversary attempts to infer whether a known individual has been part of the training data for the generative model~\citep{Yan2020GeneratingConstraints}. Evaluation can be carried out in many ways, but a naïve approach is to use DCR for each record and to predict it as a member if $\text{DCR} \leq \tau$ where $\tau$ is a fixed threshold. The reported metric can be any that evaluates the performance of the adversarial attack, but for this evaluation, the reported metrics are assumed to be precision and recall.

\textit{DOMIAS} takes a different approach to MIAs by investigating differences in densities for the synthetic data and an adversarial dataset \citep{breugel2023domias}. The method is a black-box setting assuming that an adversary has access to synthetic data, $\mathcal{S}$, and a dataset, $\mathcal{A}$, derived from the same population as the original training data, thus potentially containing instances used to train the synthetic data generation model. By producing density estimates for both $\mathcal{S}$ and $\mathcal{A}$, the method assumes overfitting in areas with higher density in $\mathcal{S}$ compared to $\mathcal{A}$. If a record in $\mathcal{A}$ is located within an overfitted area, membership is inferred. The performance is typically reported as either AUROC or accuracy.

\textit{$\epsilon$-identifiability ($\epsilon$-id)} takes a different approach and reports the proportion of synthetic records that are ``too similar'' to the true data points~\citep{Yoon2020AnonymizationADS-GAN}. Let $d$ be the minimum weighted distance within the true data, let $\hat{d}$ be the weighted distance between the true and a specific synthetic record, and let $n$ be the number of synthetic records; then $\epsilon\text{-id} = \frac{1}{n}[\mathbb{I}(\hat{d} < d)]$, where $\mathbb{I}$ is the identity function.

\textit{Identity disclosure risk (IDR)} incorporates functions like $\epsilon$-identifiability, but extends them with additional information, some of which are not readily available~\citep{ElEmam2020EvaluatingValidation}. The function takes the inverse size of the equivalence class and multiplies by: 1) an indicator variable, such as $\epsilon$-identifiability, 2) an error correction term from the literature, and 3) a binary variable indicating whether the adversary learns something new. The equivalence class is required from both the sample and the true population.

\edit{
This selection of metrics is naturally non-exhaustive, but the diversity allows for a comprehensive analysis of the capabilities of CAIR and its ability to cover a wide range of privacy metrics and notions. Numerous other metrics exist in the literature, and their omission here is solely for brevity.
}

\subsection{Findings}
\begin{figure}[tb]
    \centering
    \includegraphics[width=0.9\textwidth]{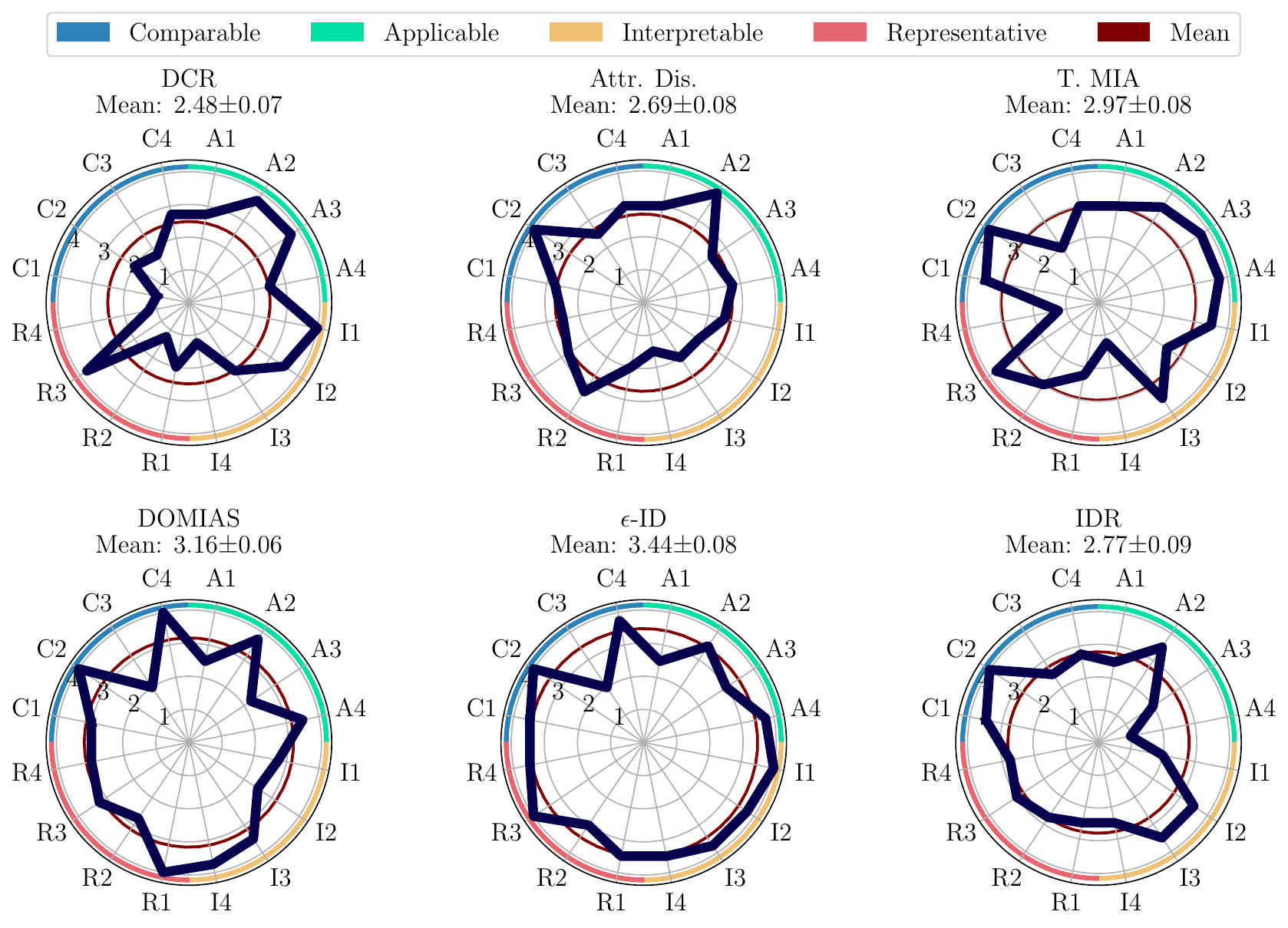}
    \caption{Radar plots of CAIR scores for the selected privacy evaluation metrics. The dimensions are indexed according to the general category and respective number and further separated into the corresponding principle denoted by the legend.}
    \label{fig:cair_scores_radar}
\end{figure}

The overall CAIR scores are given by the means with the propagated error and are presented in Figure~\ref{fig:cair_scores_radar}. Of the selected metrics, the most versatile metric for this illustrative example is $\epsilon$-identifiability with a score of $3.44 \pm 0.08$. The metric scores well in most dimensions and notably has high scores in all four main categories. In contrast, the most closely related metric to $\epsilon$-identifiability, IDR,  significantly worse and is especially penalized in the \textit{Applicability} principle for requiring information that is difficult to obtain, such as equivalence classes of the true population. Consequently, IDR gets a considerably lower CAIR score, not because the metric is less accurate but because it is difficult to employ and, thereby, not the best contender for a universal privacy metric.

Furthermore, DCR and attribute disclosure receive similar scores for a number of dimensions, but when investigating individual dimensions, it is evident that the two metrics have different downsides. Although DCR performs poorly on \textit{Comparability}, the attribute disclosure metric mainly has poor scores for \textit{Interpretability}. This exemplifies that CAIR not only allows stakeholders to rank metrics, but also allows researchers to accurately identify why they perform as they do and where improvements can be made. Interestingly, the same approach can be taken for well-performing metrics. $\epsilon$-identifiability can be improved in \textit{C3 data type agnostic} by properly accommodating mixed data types and in \textit{R1 anomalies} by improving the weighting of outliers.

DOMIAS performs the well on \textit{Representativeness} indicating that the metric generally estimates an accurate privacy level. However, the metric falls behind on \textit{Interpretability} compared to $\epsilon$-identifiability mainly due to the complex nature of density estimation and the difficulty in disseminating the underlying measurements to non-technical stakeholders. Furthermore, the metric, like all the other metrics, can be improved on \textit{C3 data type agnostic} by formally facilitating evaluation for data with both discrete and numerical data.

Threshold-based MIA and the attribute disclosure metric share many similarities, but their CAIR scores differ substantially. An examination of Figure~\ref{fig:cair_scores_radar} reveals that the scores for the \textit{Comparability} dimension are identical, while the most significant discrepancy lies in the \textit{Applicability} principle. Based on the difference in \textit{Applicability}, it can be estimated that attribute disclosure is somewhat more complex compared to threshold-based membership inference attacks.

Interestingly, CAIR score discrepancies can be utilized to identify complementary metrics that, in unison, can provide a comprehensive analysis of the privacy gain for synthetic datasets. Take, for example, DOMIAS and $\epsilon$-identifiability: they share similar scores for both \textit{Comparability} and \textit{Applicability}, while the major differences lie in \textit{Interpretability} and \textit{Representativeness}. Where DOMIAS excels at accurately representing privacy levels, $\epsilon$-identifiability is easy to interpret. Hence, a comprehensive understanding of privacy can be achieved by employing both methods as complementary evaluation methods. Although this property of CAIR is useful, the primary objective of CAIR remains to identify shortcomings and improve existing or develop new metrics that conform to the CAIR dimensions.

These findings suggest that employing a rubric to systematically scrutinize privacy evaluation metrics yields nuanced and granular insights into their efficacy as universal privacy metrics both in relation to each other and in isolation.

\section{Discussion} \label{sec:discussion}
With the increasing awareness of privacy-related issues in data management and sharing, it is more important than ever to focus on streamlining privacy evaluation, especially for maturing research fields such as synthetic tabular data generation. Many domains rely heavily on the sharing of data to advance the associated research fields. Unfortunately, the regulations put in place to protect people slow down innovation. On the other hand, regulations force institutions to consider the natural right to privacy of people. Therefore, we must work with regulators to best ensure the continuation of potentially lifesaving innovation while still maintaining the privacy of the natural person.

CAIR is an important step in the direction in which regulators and researchers collaborate and enable data sharing. Regulators are beginning to notice synthetic data as a possible substitute for true data~\citep{EuropeanComission2021ProposalACTS.}, highlighting the importance of proper evaluation of the privacy preservation capabilities of synthetic data. A natural first step is to define what constitutes a privacy metric that enables accurate privacy estimation while being sufficiently versatile to have regulations established around them. Although precision is the ultimate goal of measuring privacy levels, highly intricate and context-specific measures are difficult to regulate around making attempts at wide implementation futile. Therefore, CAIR serves as a framework for guiding metric development in a direction where precision remains an important aspect while considering outside factors that are crucial for the application of a metric.

To the best of our knowledge, CAIR is the first attempt to define principles for such measures. Therefore, this work has the potential to guide the research community in a specific direction rather than using arbitrary metrics that researchers find appropriate for their work. Furthermore, CAIR serves as guiding principles for regulators in identifying metrics and thresholds that define sufficient privacy levels to publish data. 

In summary, CAIR can help researchers, institutes, and regulators find common ground and agree on what universal privacy metrics are appropriate for synthetic tabular data.

\subsection{Composition and Application of CAIR}
One of the primary objectives of applying CAIR is to identify the strengths and weaknesses of metrics. CAIR enables researchers and other stakeholders alike to analyze how a metric functions and performs, either across the four main principles or the more granular 16 dimensions. While scoring a metric results in a single overall CAIR score, individual dimensions provide arguably more important information that can foster productive discussions and further improvements.

Still, the overall CAIR score allows a ranking of metrics according to the 16 identified dimensions. However, the CAIR score primarily reflects the universality of a privacy evaluation metric, rather than its direct quality or precision. Thus, CAIR does not guarantee that a high-scoring metric will be the most accurate metric for any given synthesis or synthetic data. This is not necessarily an issue with the definition of CAIR, but rather with the ideology of having a universal evaluation method. Therefore, we encourage the use of universal metrics alongside metrics that are more specific to the given use case. Complementing a universal metric with one or more context-specific metric(s) allows for a thorough analysis of the privacy landscape within a specific context while being comparable to studies in or outside of the given domain.

Furthermore, the ability to identify complementary privacy evaluation metrics provides added benefits by providing comprehensive coverage of the privacy landscape. However, the primary objective of CAIR is to identify single metrics that can be universally applied to synthetic tabular data generation, potentially with context-specific metrics as a complement. Thus, CAIR should be used to identify strengths and weaknesses of metrics with the intention of improving them rather than combining them. The application of CAIR in this paper suggests that these high-fidelity insights provide valuable information that might otherwise have been overlooked. Thus, CAIR provides a foundation for streamlined development of metrics.

In addition, the dimensions that have been identified as properties for good metrics are likely non-exhaustive. Consequently, future work should look at refining these properties while ensuring alignment with stakeholders. Furthermore, CAIR incentivizes researchers to consider what improvements can be made to existing privacy metrics and to design new metrics that conform to CAIR and the idea of universal evaluation metrics.

\subsection{Applicable Domain}
The CAIR principles are limited to evaluation of universal privacy metrics for synthetic tabular data and, as such, should not be applied to metrics that fall outside of that domain. Although the CAIR dimensions are designed for tabular data, the four principles (Comparability, Applicability, Interpretability, and Representativeness) may transfer well to other data modalities such as \edit{semi-structured or unstructured data, including images and natural language. However, some underlying dimensions likely require modality-specific adaptions to accurately represent challenges in the respective domains. Considering that the \textit{Interpretability} and \textit{Representativeness} principles are not concerned with the data format, these principles and their respective dimensions have a high degree of transferability to other data-modalities. Rather than focusing on the effect of the data, the focus is the expected behavior of a good and universal privacy measure.

Although the principles \textit{Comparability} and \textit{Applicability} are transferable to other modalities as overarching principles, the underlying dimensions likely require substantial adaptations to modality-specific cases. As an example, a privacy metric for image data should not be concerned with \textit{C3 Data type agnostic} as images are represented by numerical values. Similarly, while the dimension \textit{A3 Performance} may be a good fit for other modalities, the scoring definitions in the rubric, Table~\ref{tab:rubric}, may require a redefinition to better accommodate domain-specific run-time challenges.

Furthermore, some modalities may require additions or replacements of dimensions to adequately accommodate challenges. As an example on image data, the \textit{Representativeness} principle could have a dimension that denotes the metric's ability to align with human visual perception of privacy in images. Accordingly, we allow adding extra dimensions to the principles when defining CAIR for other modalities to the extent that it provides benefits. In such a case, it is important to be aware of the weighting of the overall CAIR score. Either a weighting of each dimension should be defined or each principle should have the same number of dimensions to provide equal weights, as mentioned in Section~\ref{sec:method}.
}

\edit{\subsection{Notions of Privacy}}
\edit{
Notions of privacy are changing as technological capabilities are advancing. For example, k-anonymity defines privacy as being indistinguishable from \(k-1\) other individuals \citep{Sweeney2002}, while differential privacy is concerned with the effect on an algorithmic outcome by adding or subtracting a single individual from a dataset \citep{Dwork2013ThePrivacy}. Similarly, privacy capabilities for synthetic data has various notions of privacy. Specifically, membership inference risk \citep{breugel2023domias}, attribute disclosure risk \citep{Kaur2021ApplicationData}, and similarity-based risk \citep{Guillaudeux2023Patient-centricAnalysis, Yan2020GeneratingConstraints} are common approaches to defining privacy preserving capabilities of synthetic data. However, these notions are constantly changing as new adversarial capabilities emerge.

CAIR should ideally embrace the dynamic nature of privacy evaluation. Therefore, the four principles are designed to provide a general notion of the universality of privacy evaluation metrics with the underlying dimensions denoting granular properties. Considering that CAIR already encompasses multiple notions of privacy, we argue that it is likely that the four principles will also apply to changing ideas of privacy. On the other hand, it is likely that the underlying dimensions need adaptations or extensions as the research field develops. As discussed earlier, the properties of CAIR are likely non-exhaustive and future refinements are encouraged. This is a consequence of employing a rubric for bias mitigation. The fixed definitions of the rubric values stability over flexibility of the evaluations, making it less adaptable to future development of privacy semantics. However, stability provides better bias mitigation than a vaguely defined rubric.
}

\edit{Lastly, while} some methods for introducing privacy, such as k-anonymity \citep{Sweeney2002} and differential privacy~\citep{Dwork2013ThePrivacy} allow quantification of the privacy level, they are not evaluation methods and are not limited to synthetic data. Although differential privacy provides formal privacy guarantees and can be used for synthetic data generation, setting the necessary parameters can be challenging~\citep{Jordon2021Hide-and-SeekRe-identification, Yoon2020AnonymizationADS-GAN} and can result in synthetic data that do not conform to other accepted notions of privacy, such as membership inference and attribute disclosure~\citep{Sun2023GeneratingPrivacy}. Consequently, further evaluation is needed to evaluate privacy and allow comparisons between studies. In these cases, metrics with better CAIR scores are a good choice.

\subsection{Addressing the Qualitative Nature of CAIR}
Due to the qualitative nature inherent in CAIR assessments, there exists the possibility of subjectivity, which can result in discrepancies in scores among different evaluators. However, considering that the primary objective of CAIR is to foster considerations and discussions about privacy metrics, the negative impact of subjectivity diminishes, emphasizing the usefulness and importance of even a single evaluator.

Nevertheless, when comparing CAIR scores across metrics, bias mitigation becomes increasingly important. Although the use of a rubric can mitigate bias to a certain extent, the inherent variability in interpretation calls for additional measures. An effective approach is to incorporate scores from multiple independent evaluators. Aggregating assessments from multiple perspectives minimizes the impact of individual bias and enhances the reliability of the overall assessment process. As the number of evaluators increases, the propagated error decreases, indicating scores that tend to a consensus among evaluators. The error can be interpreted as disagreement among the evaluators, which is an indicator of the level of ambiguity of a metric. In other words, additional evaluators are likely to provide additional insight into the nuances of a metric that makes cross-metric comparisons more objective.

\edit{
Disagreements in scores can occur in both a single-evaluator setting where a group acts as one evaluator and in a multi-evaluator setting. In the former case, agreements should be resolved through internal discussions that cover the technical aspects of the metric in question. These discussions contribute to a nuanced development of privacy evaluation metrics. Considering disagreements in multi-evaluator settings, discrepancies are more challenging to resolve, especially with large discrepancies. Consequently, the best approach to managing disagreements is to add more independent evaluators and consider the propagated error in the final CAIR score. If one of the evaluators significantly diverges from the others, additional evaluators will lower the error and bring the CAIR score closer to consensus. This is especially important when considering an author of a metric being one of the evaluators.
}

Furthermore, the incorporation of multiple evaluators incentivizes the research community to collaborate and collectively define good universal metrics while increasing the reliability and trustworthiness of the metrics. This collaborative approach not only enhances the reliability of the assessment but also fosters a more comprehensive understanding of the 16 CAIR dimensions leading to further improvements in CAIR.

Ultimately, researchers, regulators, and natural persons must collectively trust that any published synthetic data maintain the privacy of the individual, which can only be achieved by identifying robust generation methods and evaluating the resulting synthetic data. By evaluating the privacy-preserving capabilities with a universal privacy metric, the development of these generation methods can be accelerated by providing easy comparisons and improving reproducibility of studies.

\section{Conclusion}\label{sec:conclusion}
In this work, we have characterized four principles for good universal privacy evaluation metrics for synthetic tabular data consisting of Comparability, Applicability, Interpretability, and Representativeness (CAIR). By dividing each principle into four dimensions, we show that the CAIR rubric offers valuable granular insights into the behavior and utility of metrics, facilitating streamlined evaluation practices across various studies, allowing researchers to effectively improve privacy evaluation methodologies.

Accordingly, CAIR is a tool with multiple applications, including serving as guidelines when designing new metrics, allowing researchers to improve on existing metrics, and providing a tool for regulators to consult when defining how privacy should be evaluated for synthetic tabular data. Although CAIR relies on a qualitative assessment of metrics, it possesses the potential to act as a facilitator for the synthetic data generation community to scrutinize current and new methods to ensure robust evaluation across diverse domains. Thus, the development of the notion of privacy in synthetic data can be refined and act as a catalyst for innovation of generation methods. In addition, we expect that the CAIR principles will foster agreement among researchers and organizations on which universal privacy evaluation metrics are appropriate for synthetic tabular data.

\section*{Acknowledgments}
This study was funded by Innovation Fund Denmark in the project ``PREPARE:
Personalized Risk Estimation and Prevention of Cardiovascular Disease''.


\newpage
\appendix
\setcounter{table}{0}

\section{CAIR Scores with Explanations} \label{sec:appendix:scores}
The appendix presents explanations of the scores given to the six selected privacy evaluation metrics. Explanations are provided for the 16 dimensions separated into two tables for readability. For succinctness, only comments for Evaluator 1 are presented. Table~\ref{tab:cair_explanation1} presents the first two metrics: $\epsilon$-identifiability, Identity Disclosure Risk (IDR), Table~\ref{tab:cair_explanation2} presents Distance to Closest Record (DCR) and attribute disclosure (Attr. Dis.), and Table~\ref{tab:cair_explanation3} presents threshold-based membership inference attack (T. MIA). and DOMIAS. The CAIR scores for each metric are the mean of dimension means across the evaluators with the propagated error as described in Figure~\ref{fig:cair_evaluation_flow}.

\begin{footnotesize}
\begin{longtable}
{L{0.15\textwidth}C{0.06\textwidth}L{0.3\textwidth}C{0.06\textwidth}L{0.3\textwidth}}
\caption{Presentation of CAIR scores for $\epsilon$-identifiability and Identity Disclosure Risk (IDR). The scores by each evaluator are represented as $E_1/E_2$ for Evaluators 1 and 2, respectively.} \label{tab:cair_explanation1}\\
\toprule
 \textbf{Dimensions} & $\mathbf{E_1/E_2}$ & \textbf{$\epsilon$-Identifiability} & $\mathbf{E_1/E_2}$ & \textbf{IDR}\\
 \cmidrule(rl){1-1} \cmidrule(rl){2-3} \cmidrule(rl){4-5} 
\endfirsthead
\toprule
 \textbf{Dimensions} & $\mathbf{E_1/E_2}$ & \textbf{$\epsilon$-Identifiability} & $\mathbf{E_1/E_2}$ & \textbf{IDR} \\
 \cmidrule(rl){1-1} \cmidrule(rl){2-3} \cmidrule(rl){4-5}
\endhead

    \multicolumn{5}{l}{\textbf{Comparability}} \\\\
    \textbf{C1 \newline Scale} & 
        3/4 & Mostly invariant, but relies on distance functions without normalization. &
        3/4 & Mostly invariant, but relies on distance functions without normalization. \\ \cmidrule{2-5}

    \textbf{C2 \newline Metric bounds} & 
        4/4 & Bounds are $[0, 1]$. &
        4/4 & Bounds are $[0, 1]$. \\ \cmidrule{2-5}

    \textbf{C3 \newline Data type agnostic} &
        2/2 & Depends on the choice of distance function. Authors suggest Euclidean distance, which performs poorly on categorical data. Consequences can potentially be severe. Remediations can be cumbersome. &
        2/3 & No clear definition of distance function is given. Consequences can potentially be severe. Remediations can be cumbersome. \\ \cmidrule{2-5}

    \textbf{C4 \newline Cross-domain relevance} &
    4/3.5 & Invariant to the domain. &
        3/2.5 & The connection between training data and the population is not necessarily relevant to all domains. \\ \midrule

    \multicolumn{5}{l}{\textbf{Applicability}} \\\\
    \textbf{A1 \newline Heteroge-neity} & 
        3/2 & Depends on the choice of distance function(s). The authors suggest Euclidean distance, but explicitly leave the choice of distance function to the user. & 
        3/2 & Depends on the choice of distance function(s), but no clear definition is provided.\\ \cmidrule{2-5}

    \textbf{A2 \newline Diverse generation methods} &
        4/3 & Invariant to the generation method. &
        4/3 & Invariant to the generation method. \\ \cmidrule{2-5}

    \textbf{\edit{A3 \newline Performance}} &
        3/3 & Computes the pairwise distances within the true data and between the true and synthetic data $\mathcal{O}(2n^2)$. &
        2/2 & Runs in at least $\mathcal{O}(N^2 + n^2)$ where $n$ is the sample size and $N$ is population size. $N$ can potentially be extremely large and expensive to estimate. \\ \cmidrule{2-5}

    \textbf{A4 \newline Implemen-tation} &
        4/3.5 & An implementation is provided and the metric is well-documented. &
        1/1 & The metric requires information from the population, which is difficult to obtain or estimate. The error correction parameter $\lambda$ is external information from the literature. The parameter $R$ "learning something new" is ambiguous and not well-defined and can be challenging to compute. \\ \midrule 

    \multicolumn{5}{l}{\textbf{Interpretability}} \\\\
    \textbf{I1 \newline Explain-ability} &
        4/4 & Measures the proportion of synthetic data points that are "different enough" from true data points. Simple percentage of how many records are at risk. &
        2/2 & Requires knowledge on data errors, sampling, and populations, but can be communicated to someone with some degree of prior knowledge. \\ \cmidrule{2-5}

    \textbf{I2 \newline Understand-ability} & 
        4/3.5 & The output is the risk in percentage, which is a common concept for laypeople. &
        4/3 & The output is the risk in percentage, which is a common concept for laypeople. \\ \cmidrule{2-5}

    \textbf{I3 \newline Visualization} &
        4/3.5 & The output is a single value that can easily be visualized. &
        4/3 & The output is a single value that can easily be visualized. \\ \cmidrule{2-5}

    \textbf{I4 \newline Granularity} &
        4/3 & The output is linear with a single value as the output. &
        3/2 & The output is linear with a single value as the output.  It is unclear how changes to the output are due to the synthetic data or the population. \\ \midrule 

    \multicolumn{5}{l}{\textbf{Representativeness}} \\\\
    \textbf{R1 \newline Anomalies} & 
        3/4 & The metric takes all records into account, but the weighting is not definitively defined. The authors suggest using inverse column-wise entropy, but other weightings may be more appropriate. &
        3/2 & The metric is weighted with the inverse size of the equivalence class, but is unclear how appropriate such a weighting is. \\ \cmidrule{2-5}

    \textbf{R2 \newline Coverage} & 
        3/4 & Depends on the choice of weightings. The authors suggest inverse column-wise entropy but explicitly leave the choice for the user. &
        3/2.5 & Depending on how the population is computed, the result may not be accurate for all classes. \\ \cmidrule{2-5}

    \textbf{R3 \newline Reproducibility} &
        4/4 & Always produces the same output for the same input. &
        3/3 & Depends on how the population is estimated and the definition of "learning something new". Without clear definitions, the results may not be reproducible in all instances. \\ \cmidrule{2-5}

    \textbf{R4 \newline Precision} &
        3/4 & Slightly overestimates privacy violations as the same synthetic data record can pose a violation for an arbitrary number of real records. &
        3/2.5 & Slightly overestimates privacy violations as the same synthetic data record can pose a violation for an arbitrary number of real records.  \\ \midrule
    
    \textbf{CAIR} & \multicolumn{2}{c}{$\mathbf{3.44 \pm 0.08}$} & \multicolumn{2}{c}{$\mathbf{2.77 \pm 0.09}$}\\
    \bottomrule
\end{longtable}
\end{footnotesize}

\begin{footnotesize}
\begin{longtable}
{L{0.15\textwidth}C{0.06\textwidth}L{0.3\textwidth}C{0.06\textwidth}L{0.3\textwidth}}
\caption{Presentation of CAIR scores for Distance to Closest Record (DCR) an attribute disclosure (Attr. Dis.). The scores by each evaluator are represented as $E_1/E_2$ for Evaluators 1 and 2, respectively.} \label{tab:cair_explanation2}\\
\toprule
 \textbf{Dimensions} & $\mathbf{E_1/E_2}$ & \textbf{DCR} & $\mathbf{E_1/E_2}$ & \textbf{Attr. Dis.} \\
 \cmidrule(rl){1-1} \cmidrule(rl){2-3} \cmidrule(rl){4-5} 
\endfirsthead
\toprule
 \textbf{Dimensions} & $\mathbf{E_1/E_2}$ & \textbf{DCR} & $\mathbf{E_1/E_2}$ & \textbf{Attr. Dis.}  \\
 \cmidrule(rl){1-1} \cmidrule(rl){2-3} \cmidrule(rl){4-5}
\endhead
    \multicolumn{5}{l}{\textbf{Comparability}} \\\\
    \textbf{C1 \newline Scale} & 
        1/1 & Completely dependent on the scale of the data. & 
        3/2.5 & The metric is largely invariant to the data scale, but normalization is recommended. \\ \cmidrule{2-5}

    \textbf{C2 \newline Metric bounds} & 
        2/2 & The metric only has a lower bound $[0, \infty)$. &
        4/4 & Bounds for both outputs are well defined: \newline Precision: $[0,1]$ \newline Recall: $[0,1]$ \\ \cmidrule{2-5}

    \textbf{C3 \newline Data type agnostic} & 
        2/1.5 & Depends entirely on the choice of distance function, but without considering how functions are applied, the results can be significantly affected. &
        2/3 & Depends on the choice of distance function, but without considering how functions are applied, the results can be significantly affected.  \\ \cmidrule{2-5}

    \textbf{C4 \newline Cross-domain relevance} &
        2.5/3	& Considering some domains use data with a large number of attributes, DCR would not be appropriate for such domains as the distance between records will grow with the number of attributes. For example, in genomics. &
        3/3 & Relevant for most domains as long as the thresholds correspond to the data. \\ \midrule

    \multicolumn{5}{l}{\textbf{Applicability}} \\\\
    \textbf{A1 \newline Heteroge-neity} & 
        3/2.5 & Depends on the choice of distance function, which is not for defined for the metric. &
        3/3 & Depends on the choice of distance function, which is not defined for the metric. \\ \cmidrule{2-5}

    \textbf{A2 \newline Diverse generation methods} &
        4/3.5 & Invariant to generation method. &
        4/4 & Invariant to generation method.  \\ \cmidrule{2-5}

    \textbf{\edit{A3 \newline Performance}} &
        4/3.5 & Only uses the pairwise distances. $\mathcal{O}(n^2)$. &
        3/2 & $\mathcal{O}(kn^2)$ where k is the number of different levels of knowledge for the adversaries that are tested. \\ \cmidrule{2-5}

    \textbf{A4 \newline Implemen-tation} &
        3/2 & The metric is very simple and easy to implement, but poses some ambiguity. &
        3/2.5 & There is some ambiguity in how predictions are performed and in what order. Additionally, the number of adversarial knowledge levels can vary. \\ \midrule

    \multicolumn{5}{l}{\textbf{Interpretability}} \\\\
    \textbf{I1 \newline Explain-ability} &
        4/4 & Very simple and easy to explain. &
        2/3 & A layperson is unlikely to be familiar with the concepts of prediction and adversarial knowledge. Substantial explanations are required. Some background knowledge is necessary. \\ \cmidrule{2-5}

    \textbf{I2 \newline Understand-ability} & 
        3/4 & The concept of distance functions is not necessarily known by laypeople, but can easily be explained. &
        2/2 & The output is precision and recall, which are unfamiliar concepts to laypeople. Without some background knowledge in statistics or data management, it can be difficult to understand. \\ \cmidrule{2-5}

    \textbf{I3 \newline Visualization} &
        2/3 & Comparisons are difficult considering that the metric has no upper bound and the scale is not the same across datasets. &
        2/2 & Visual comparisons become challenging when multiple adversarial levels are being tested unless aggregate data are used (in which case the visualization is no longer completely accurate). \\ \cmidrule{2-5}

    \textbf{I4 \newline Granularity} &
        1.5/1 & Privacy levels can be differentiated, but due to the effect of the scale of the data, some privacy violations may be overlooked. The value only makes sense in comparison to other results. &
        1/2 & The output potentially consists of many sets of values, but at least two values. \\ \midrule

    \multicolumn{5}{l}{\textbf{Representativeness}} \\\\
    \textbf{R1 \newline Anomalies} & 
        2/2 & All observations are treated equally. &
        2/2 & There is no specific weighting for data points. A potential weighting should be different for various levels of adversarial knowledge. \\ \cmidrule{2-5}

    \textbf{R2 \newline Coverage} & 
        1/1.5 & DCR is only ever representative of small subsets of the data, whether the mean, median, or individual DCRs are considered. &
        3/3.5 & The performance of the predictive attack represents most points, but without considering outliers, some might have overlooked privacy violations.  \\ \cmidrule{2-5}

    \textbf{R3 \newline Reproducibility} &
        4/3.5 & Always produces the same output for the same input. &
        3/2.5 & Depends on the implementation and how the prior knowledge is selected. \\ \cmidrule{2-5}

    \textbf{R4 \newline Precision} &
        1/1.5 & When using the median or mean DCR, the privacy level is significantly underestimated. &
        2/3 & Somewhat favors inliers. \\ \midrule    
    
    \textbf{CAIR} & \multicolumn{2}{c}{$\mathbf{2.48 \pm 0.07}$} & \multicolumn{2}{c}{$\mathbf{2.69 \pm 0.08}$}\\
    \bottomrule
\end{longtable}
\end{footnotesize}

\newpage
\begin{footnotesize}
\begin{longtable}
{L{0.15\textwidth}C{0.06\textwidth}L{0.3\textwidth}C{0.06\textwidth}L{0.3\textwidth}}
\caption{Presentation of CAIR scores for threshold-based MIA (T. MIA) and DOMIAS. The scores by each evaluator are represented as $E_1/E_2$ for Evaluators 1 and 2, respectively.} \label{tab:cair_explanation3}\\
\toprule
 \textbf{Dimensions} & $\mathbf{E_1/E_2}$ & \textbf{T. MIA} & $\mathbf{E_1/E_2}$ & \textbf{DOMIAS} \\
 \cmidrule(rl){1-1} \cmidrule(rl){2-3} \cmidrule(rl){4-5} 
\endfirsthead
\toprule
 \textbf{Dimensions} & $\mathbf{E_1/E_2}$ & \textbf{T. MIA} & $\mathbf{E_1/E_2}$ & \textbf{DOMIAS} \\
 \cmidrule(rl){1-1} \cmidrule(rl){2-3} \cmidrule(rl){4-5}
\endhead
        \multicolumn{5}{l}{\textbf{Comparability}} \\\\
    \textbf{C1 \newline Scale} & 
        3/4 & The metric is largely invariant to the data scale, but normalization is recommended. &
        3/3 & Mostly invariant, but relies on density estimation methods that may require normalization. \\ \cmidrule{2-5}

    \textbf{C2 \newline Metric bounds} & 
        4/4 & Bounds for both outputs are well defined: \newline Precision: $[0,1]$ \newline Recall: $[0,1]$ &
        4/4 & Bounds are [0, 1] \\ \cmidrule{2-5}

    \textbf{C3 \newline Data type agnostic} &
        2/2 & Depends on the choice of distance function, but without considering how functions are applied, the results can be significantly affected. &
        2/2 & Depends on the density estimation method. The authors suggest BNAF or Gaussian KDE, which do not inherently work with categorical data. \\ \cmidrule{2-5}

    \textbf{C4 \newline Cross-domain relevance} &
        3/3 & Relevant for most domains as long as the threshold is appropriate to the domain. & 
        4/4 & Invariant to the domain.\\ \midrule

    \multicolumn{5}{l}{\textbf{Applicability}} \\\\
    \textbf{A1 \newline Heteroge-neity} & 
        3/3 & Depends on the choice of distance function, which is not defined for the metric. &
        3/2 & Depends on the choice of density estimation og possible underlying distance function. \\ \cmidrule{2-5}

    \textbf{A2 \newline Diverse generation methods} &
        4/3 & Invariant to the generation method.& 
        4/3.5 & Invariant to the generation method. \\ \cmidrule{2-5}

    \textbf{\edit{A3 \newline Performance}} &
        4/3.5 & Computes pairwise distances $\mathcal{O}(n^2)$. & 
        2.5/2 & Depends on the choice of density estimation and possible underlying distance function. BNAF is computationally expensive while Gaussian KDE is $\mathcal{O}(n^2)$. Accordingly, DOMIAS is at best $\mathcal{O}(2n^2)$.\\ \cmidrule{2-5}

    \textbf{A4 \newline Implementation} &
        4/3.5 & Easy to compute and implement. &
        3.5/3.5 & An implementation is provided by the authors, but the choice of density estimator is ambiguous. \\ \midrule

    \multicolumn{5}{l}{\textbf{Interpretability}} \\\\
    \textbf{I1 \newline Explain-ability} &
        3/4 & The concept can be explained as ``does this synthetic data point look too similar to a true record?''. However, explaining how similarity relates to membership inference can be somewhat difficult to understand for laypeople. &
        2.5/3 & Understanding the use of density estimation likely requires some domain knowledge. \\ \cmidrule{2-5}

    \textbf{I2 \newline Understand-ability} & 
        2/3 & The output is precision and recall for an adversarial attack, which are unfamiliar concepts to laypeople. Without some background knowledge in statistics or data management, it can be difficult to understand. &
        2.5/2.5 & Depends on the choice of performance metric. Accuracy is well-understood, while AUROC requires some explanation. \\ \cmidrule{2-5}

    \textbf{I3 \newline Visualization} &
        4/3 & Even though the output consists of two values, they can easily be represented visually. &
        4/3 & The output is a single value that is easy to visualize. \\ \cmidrule{2-5}

    \textbf{I4 \newline Granularity} &
        1/1.5 & The output consists of two values, making differentiation between privacy levels quite challenging. & 
        4/3.5 & The output is linear with a single value as the output. \\ \midrule

    \multicolumn{5}{l}{\textbf{Representativeness}} \\
    \textbf{R1 \newline Anomalies} & 
        2/2.5 & The threshold is the same for all data points. &
        4/4 & Differences in density estimations around outliers are likely greater than for inliers. \\ \cmidrule{2-5}

    \textbf{R2 \newline Coverage} &
        3/3 & The performance of the predictive attack represents most points, but without considering outliers some might have overlooked privacy violations. & 
        3/2.5 & The coverage can be affected by the number of features and the corresponding sample size. \\ \cmidrule{2-5}

    \textbf{R3 \newline Reproducibility} &
        4/3.5 & Always produces the same output for the same input. &
        3/3.5 & Depends on the choice of density estimation method. \\ \cmidrule{2-5}

    \textbf{R4 \newline Precision} &
        1/1.5 & Systematically favors inliers by having a fixed similarity threshold. &
        3/3 & Potentially underestimates the privacy level. \\ \midrule
    
    \textbf{CAIR} & \multicolumn{2}{c}{$\mathbf{2.97 \pm 0.08}$} & \multicolumn{2}{c}{$\mathbf{3.16 \pm 0.06}$}\\
    \bottomrule
\end{longtable}
\end{footnotesize}

\end{document}